\documentclass[11pt]{article}

\usepackage[final]{acl}

\usepackage{times}
\usepackage{latexsym}

\usepackage[T1]{fontenc}

\usepackage[utf8]{inputenc}

\usepackage{microtype}

\usepackage{inconsolata}

\usepackage{graphicx}

\usepackage{amsmath}
\usepackage{booktabs}
\usepackage{colortbl}
\usepackage{multirow}
\usepackage{wrapfig}
\usepackage{makecell}
\usepackage[accsupp]{axessibility}
\usepackage{hyperref}
\usepackage{orcidlink}
\usepackage{xspace}

\newcommand{\eg}{e.g.,\xspace}

\title{HAP: Head-Adaptive Visual Token Pruning via Cross-Modal Alignment}

\author{
  Yuanhao Sun$^{1}$, Huawei Ji$^{1}$, Yuan Jin$^{1}$, Cheng Deng$^{2}$, Luoyi Fu$^{1*}$, Xinbing Wang$^{1}$ \\
  $^{1}$Shanghai Jiao Tong University, Shanghai, China \\
  $^{2}$University of Edinburgh, Edinburgh, UK \\
  \texttt{\{h\_iden, sjtua3365981, lemon0703, yiluofu, xwang8\}@sjtu.edu.cn}
}

\begin{document}
\maketitle
\let\thefootnote\relax\footnotetext{*Corresponding author}

\begin{abstract}
Recent Vision-Language Models encode high-resolution images into long visual token sequences, incurring prohibitive prefill costs. To compress them, existing methods score each visual token by averaging text-to-visual attention uniformly across all heads, which assumes every head matches the query. However, our empirical analysis shows that misaligned heads dominate the average, amplifying background tokens and drowning out fine-grained cues.

To address this, we propose PAQ (Prompt-Grounded Attention Quality), a metric quantifying how well each head aligns the prompt with image regions. Built on PAQ, our pruning proceeds in three stages. Given a target FLOPs budget, we first partition the transformer layers into groups and allocate a visual token budget to each. Within each group, we then aggregate per-head attention maps via PAQ-weighted softmax into a group-level matrix. Finally, we score visual tokens by this matrix's magnitude and retain the allocated budget per group. By weighting heads with PAQ, our method scores tokens by attention signals that more faithfully reflect prompt relevance, rather than diluting them through uniform averaging.

Across 18 benchmarks, our method delivers state-of-the-art trade-offs. Specifically, on LLaVA-1.5-7B (9 tasks), retaining only \textbf{5.6\%} tokens preserves \textbf{99.1\%} of the original performance, surpassing the strongest baseline AutoPrune by 4.2 points. Code is available in https://github.com/baokou-fw2/HAP.

\end{abstract}

\section{Introduction}
\label{sec:intro}

Recent advances in Vision-Language Models (VLMs)~\cite{liu2023visual,lu2026internvl,Qwen2.5-VL} have achieved remarkable progress in multimodal reasoning and understanding. To capture fine-grained details, state-of-the-art VLMs increasingly adopt high-resolution visual encoders that produce long sequences of visual tokens.
This visual token explosion imposes substantial computational and memory overhead during the autoregressive prefill stage, creating a critical bottleneck for real-time deployment and limiting the scalability of VLMs in resource-constrained environments.

\begin{figure}[t]
\centering
\includegraphics[width=1.0\linewidth]{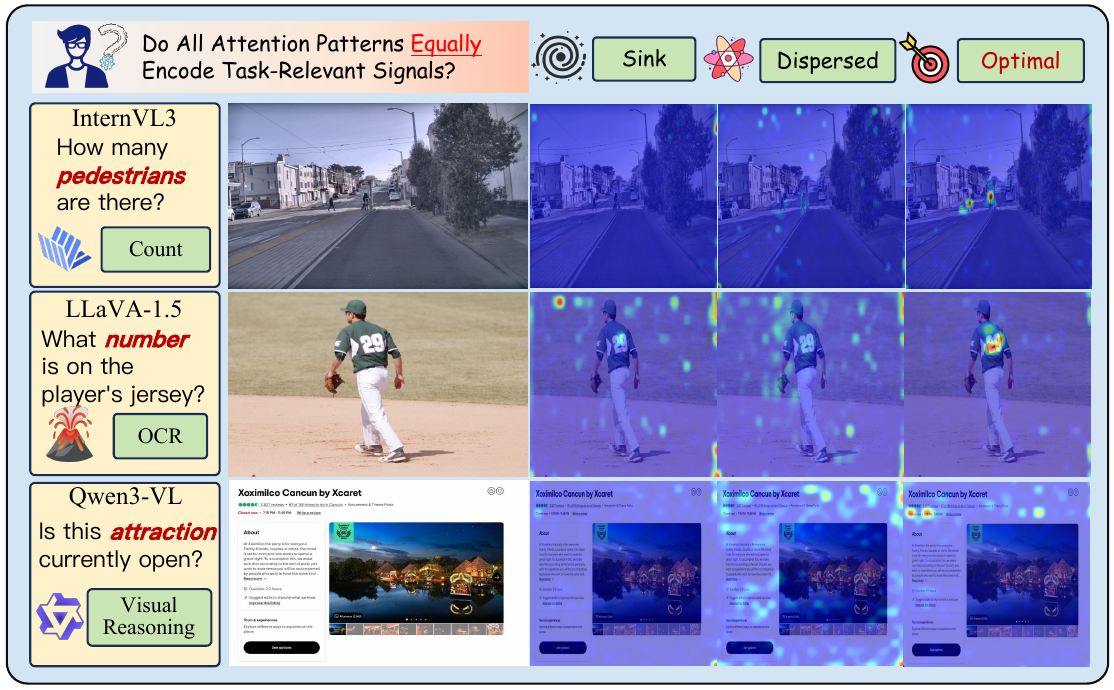}
\caption{\textbf{Head-level cross-modal attention patterns.} Layer 2 
attention maps reveal \emph{sink}, \emph{optimal}, and 
\emph{dispersed} cross-modal patterns across three VLMs.}
\label{fig0}
\vspace{-5mm}
\end{figure}

To mitigate this prefill burden, existing methods exploit cross-modal attention statistics to rank and prune visual tokens. These weights are already computed in the forward pass and naturally indicate where the model focuses under the text prompt. For instance, SparseVLM~\cite{zhang2024sparsevlm} identifies informative text tokens as raters and accumulates their attention mass over visual tokens to produce a layer-wise importance score. PDrop~\cite{xing2024pyramiddrop} aggregates attention scores uniformly across all heads within each target layer and progressively discards low-ranking tokens following a pyramid schedule. AutoPrune~\cite{wang2026each} further derives mutual information from attention distributions to guide layer-wise budget allocation.

\begin{figure}[t]
\centering
\includegraphics[width=\linewidth]{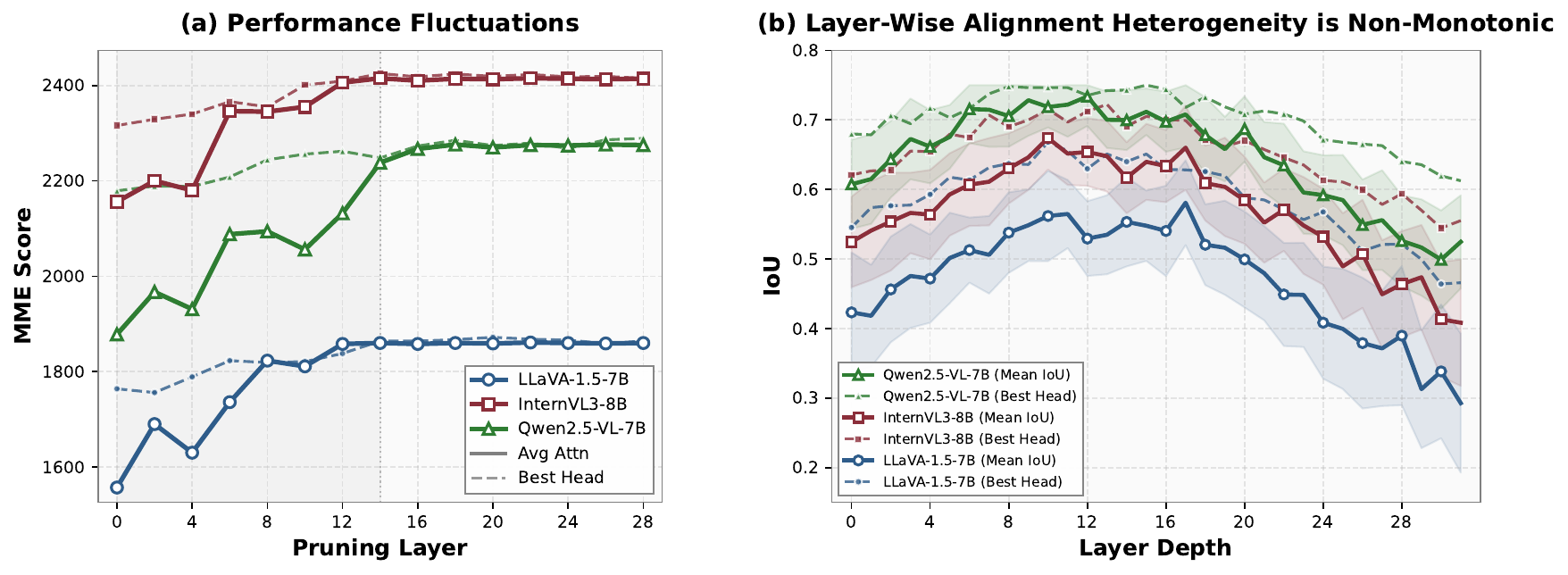}
\caption{(a)~MME performance when retaining 25\% of visual tokens at different layers across three VLMs. (b)~RefCOCO IoU when retaining 50\% of visual tokens at different layers. In both panels, solid curves denote average attention mass and dashed curves denote the best single head; average aggregation consistently trails the optimal head.}
\label{fig1}
\vspace{-5mm}
\end{figure}

Despite their varied designs, these methods share a critical simplification. They implicitly assume that all attention heads within a layer exhibit the same degree of focus on prompt-relevant regions. Consequently, they average attention scores indiscriminately across all heads. However, our empirical analysis in Fig.~\ref{fig0} reveals attention heads split into three distinct modes (sink, dispersed, and optimal) with starkly different spatial focus.
Sink heads collapse to fixed regions regardless of the text prompt. Dispersed heads spread attention broadly without clear prompt correspondence. Only optimal heads concentrate on task-relevant regions that shift with the query. This reveals a critical yet overlooked property. The image regions that the model attends to under text guidance vary drastically across heads. 
We term this property \emph{prompt-grounded attention}. 
Fig.~\ref{fig1} makes this cost concrete, showing that tokens selected by uniform head averaging consistently fall short of those selected by the single best-aligned head, both in task accuracy and in spatial localization.

To distinguish reliable prompt-grounded attention from spurious noise, we propose the \textbf{Prompt-Grounded Attention Quality (PAQ)} score. 
Based on the uncertainty coefficient~\cite{theil1970estimation}, PAQ is computed purely from post-softmax intrinsic attention statistics. 
It measures how genuinely an attention head's visual distribution is driven by the text prompt. 
A high PAQ indicates concentrated attention on semantically prompt-relevant regions, while a low PAQ reflects attention-sink or dispersed patterns that are unreliable for pruning.

Building upon PAQ, we introduce \textbf{HAP}, a head-adaptive visual token pruning framework that requires no per-task hyper-parameter tuning. Given a target FLOPs budget, HAP organizes transformer layers into groups and allocates visual token budgets following a geometric pyramid structure, progressively reducing tokens from lower to higher layers. Within each layer, HAP evaluates every attention head with PAQ. Heads with high PAQ, which concentrate on prompt-relevant regions, receive greater weight in the scoring. Heads with low PAQ are effectively softly suppressed. This produces a quality-aware layer-level score. HAP then hierarchically merges these layer scores across the group under the same PAQ weighting to obtain the final token importance ranking. This two-stage design replaces uniform head averaging with a principled mechanism that scores visual tokens by their true prompt relevance.

We validate the effectiveness of our method across 18 benchmarks and five model architectures, including LLaVA, InternVL, Qwen-VL and Deepseek-VL. In terms of accuracy, retaining only 5.6\% of the visual tokens, our model still preserves 99.1\% of the original performance, outperforming the strongest baseline AutoPrune by 4.2 points. In terms of efficiency, it reduces KV-cache usage by 80\% and achieves a 2.82× inference speedup. Our contributions are summarized as follows:
\begin{itemize}
    \item We empirically reveal diverse cross-modal alignment patterns across attention heads. We propose PAQ, a metric that discriminates reliable alignment from attention-sink or dispersed patterns in a single score.
    \item We propose HAP, a training-free pruning strategy whose only user-specified input is the target FLOPs budget. HAP replaces coarse head averaging with PAQ-guided softmax aggregation, so that visual tokens are scored by attention signals that better reflect prompt relevance.
    \item We evaluate HAP on five representative VLM architectures across 18 benchmarks. Results show that HAP achieves state-of-the-art performance-efficiency trade-offs, validating the strong generalizability of our approach.
\end{itemize}

\section{Related Work}
\paragraph{Vision-Language Models.}
Recent large vision-language models (VLMs) have achieved remarkable multimodal reasoning by integrating high-resolution visual encoders with LLMs~\cite{hurst2024gpt, liu2023visual, guo2025deepseek, Qwen2.5-VL,sun2026encore}.
However, these encoders generate dense visual token sequences (ranging from hundreds to thousands per image), creating substantial computational and memory bottlenecks during prefill~\cite{liu2024llavanext, chen2024internvl}.
This \emph{visual token explosion} critically hinders efficient deployment of modern VLMs.

\paragraph{Visual Token Pruning for VLMs.}
Visual token pruning approaches fall into training-based and training-free paradigms.
Training-based methods redesign the visual encoder or inject learnable compression modules~\cite{xing2024pyramiddrop, cha2024honeybee, zhang2025llava}, requiring expensive retraining and lacking scalability across diverse backbones.
Training-free methods avoid retraining by leveraging unimodal visual attention or cross-modal attention matrices.
Visual-modal approaches~\cite{arif2025hired, zhang2025beyond} select tokens based on visual saliency but inherently lack textual guidance, often discarding task-relevant background.
Cross-modal methods~\cite{zhang2024sparsevlm, wang2026each, zhang2025vscan} exploit vision-language interactions but typically aggregate attention heads uniformly and apply pruning at pre-defined layers, failing to exploit the heterogeneous cross-modal alignment quality across different model components.

\paragraph{Attention Head Heterogeneity.}
In text-only LLMs, attention heads are known to be heterogeneous: attention sinks concentrate probability mass on delimiter or initial tokens~\cite{xiao2023efficient}, and individual heads exhibit functional specialization~\cite{zheng2024attention}.
Related phenomena also appear in VLMs, but with qualitatively different semantics: visual sinks fixate on image backgrounds or edges rather than textual delimiters~\cite{kang2025see}, and VLM heads follow heterogeneous routing behaviors that alternate between visual-sink and dispersed patterns~\cite{luo2025tosink}.
This structural difference means that LLM-oriented treatments of sink heads do not transfer directly to cross-modal pruning.
HAP builds on this distinction: instead of treating head heterogeneity as noise to be averaged out, PAQ exploits it as a reliability signal and actively down-weights misaligned heads, so that only prompt-relevant heads dominate the pruning signal (see Appendix~\ref{app:llm_vlm} for a detailed discussion).

\section{Method}
\label{sec:method}

\begin{figure}[t]
\centering
\includegraphics[width=\linewidth]{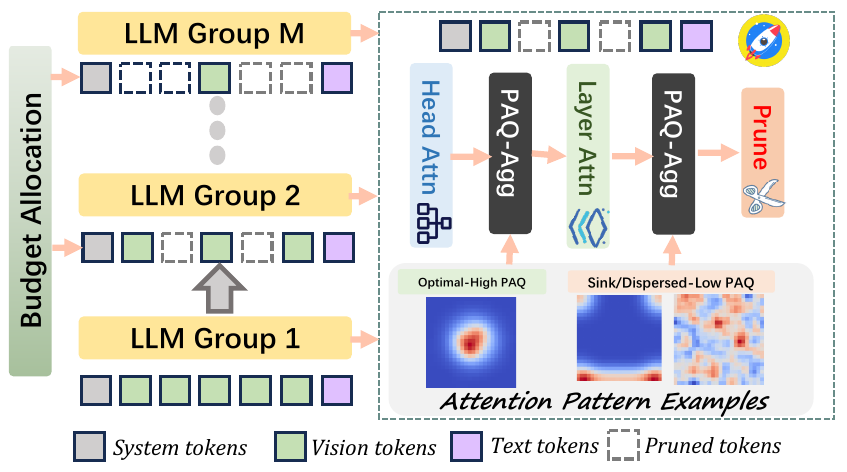}
\caption{\textbf{Overview of HAP.} 
Left: FLOPs-driven budget allocation progressively reduces visual tokens across $M$ layer groups. 
Right: unified PAQ-softmax aggregation. Head attention maps are softly fused within each layer and then aggregated within each group, yielding a reliable score for visual token pruning.}
\label{fig2}
\vspace{-5mm}
\end{figure}

In this section, we formalize the VLM pruning objective and identify two critical limitations of strategies: uniform head aggregation and fixed-layer pruning (Sec.~\ref{sec:preliminary}). We then propose the \textbf{Prompt-Grounded Attention Quality} (PAQ) score to discriminate reliable cross-modal alignment from misaligned and dispersed patterns (Sec.~\ref{sec:PAQ}). Building upon PAQ, we introduce \textbf{HAP} (Sec.~\ref{sec:HAP}). Finally, we conclude with theoretical analysis of computational efficiency and system compatibility (Sec.~\ref{sec:analysis}). An overview is shown in Fig.~\ref{fig2}.

\subsection{Rethinking Token Pruning}
\label{sec:preliminary}

VLMs generate textual responses conditioned on visual inputs and text prompts. An image is encoded into visual tokens $\mathbf{V} \in \mathbb{R}^{N_v \times D}$, accompanied by a system prompt $\mathbf{S} \in \mathbb{R}^{N_s \times D}$ and a user query $\mathbf{T} \in \mathbb{R}^{N_t \times D}$. The model auto-regressively generates a response $\mathbf{R}$ by modeling $p(\mathbf{R} \mid \mathbf{S}, \mathbf{V}, \mathbf{T})$.

Token pruning is formalized as a query-conditioned relevance ranking problem at the prefill stage, retaining only the visual tokens most relevant to the text query. Previous methods~\cite{zhang2024sparsevlm,wang2025dymudynamicmergingvirtual} compute a cross-modal score $s(\mathbf{v}_i)$ for each token and retain the top-$K$:
\begin{equation}
    \mathbf{V}' = \{\mathbf{v}_i \mid s(\mathbf{v}_i) \in \mathrm{top}\text{-}K(\{s(\mathbf{v}_j)\}_{j=1}^{N_v})\},
    \label{eq:prune_objective}
\end{equation}
where $K$ is a pre-defined budget. Concretely, these methods aggregate attention scores uniformly across all $H$ heads in a fixed layer $k$:
\vspace{-3.5mm}
\begin{equation}
    s(\mathbf{v}_i) = \frac{1}{H}\sum_{h=1}^H \mathbf{A}_h^k(\mathbf{v}_i, \mathbf{T}),
    \label{eq:naive_score}
\end{equation}
where $\mathbf{A}_h^k(\mathbf{v}_i, \mathbf{T})$ measures the total attention mass from text to visual token $\mathbf{v}_i$ in head $h$.

However, Eq.~\eqref{eq:naive_score} assumes all attention heads contribute equally to prompt-grounded visual focus, which fails in practice. As Fig.~\ref{fig0} illustrates, heads within a single layer exhibit heterogeneous patterns (sink, dispersed, and optimal), rendering uniform aggregation suboptimal. These patterns also manifest across layers and architectures.

We conduct two complementary experiments to quantify the cost of ignoring this head-level variance. 
In Fig.~\ref{fig1}a, we retain 25\% of visual tokens at each layer independently and measure MME~\cite{zhang2024mme} performance across three VLMs. 
In Fig.~\ref{fig1}b, we retain 50\% of tokens at each layer on RefCOCO~\cite{kazemzadeh2014referitgame} and measure IoU. 
Both figures consistently reveal the same gap: \textbf{visual tokens selected by the average attention mass consistently trail those selected by the best single head} in both task performance and spatial precision. 
Uniform aggregation therefore dilutes the fine-grained signals from optimal heads while amplifying noise from sink and dispersed ones.

Together, the layer-wise fluctuation and head-wise gap demonstrate that token pruning requires selective exploitation of prompt-grounded attention rather than uniform aggregation.

\subsection{Prompt-Grounded Attention Quality}
\label{sec:PAQ}
To select heads that encode reliable visual focus driven by the text prompt, we define the \textbf{Prompt-Grounded Attention Quality (PAQ)} score. 
Following the \emph{uncertainty coefficient}~\cite{theil1970estimation}, PAQ is formulated as:
\begin{equation}
    \mathrm{PAQ} = \frac{I(\mathbf{V};\mathbf{T})}{H(\mathbf{V})},
    \label{eq:PAQ}
\end{equation}
where $I(\mathbf{V};\mathbf{T})$ measures how much the visual distribution is driven by the text query, and $H(\mathbf{V})$ measures total visual uncertainty. This ratio, bounded between 0 and 1, quantifies the proportion of visual distribution explained by the query.

We compute PAQ purely from intrinsic attention statistics, without extra parameters. Following prior work~\cite{vaswani2017attention,wang2026each}, we consider the cross-modal attention map that encodes text-to-vision alignment. Let $\alpha_{ji}$ denote the normalized attention weight from the $j$-th text token $t_j$ to the $i$-th visual token $v_i$ in a given head. We interpret this cross-modal weight as a soft relevance assignment $\hat{p}(v_i \mid t_j) = \alpha_{ji}$, while adopting a uniform text prior $p(t_j)=1/N_t$.
This yields the joint distribution $p(v_i, t_j) = \frac{1}{N_t}\alpha_{ji}$ and marginal $p(v_i) = \sum_j p(v_i, t_j)$. 
Substituting these into Eq.~\eqref{eq:PAQ} gives the PAQ score:
\begin{equation}
\begin{split}
    I(\mathbf{V};\mathbf{T}) &= \sum_{i,j} p(v_i,t_j)\log\frac{p(v_i,t_j)}{p(v_i)p(t_j)}, \\
    H(\mathbf{V}) &= -\sum_i p(v_i)\log p(v_i).
\end{split}
\label{eq:PAQ_full}
\end{equation}

Consequently, the three failure modes from Fig.~\ref{fig0} map directly onto this formulation. For both \textbf{sink} and \textbf{dispersed} heads, the attention weights $\alpha_{ji}$ are invariant to the text token $t_j$. 
In a sink head, every text token attends to the same small set of visual tokens. In a dispersed head, every text token spreads attention uniformly over all visual tokens. 
In either case, the conditional probability $\hat{p}(v_i \mid t_j)=\hat{p}(v_i)$ for all $j$, meaning the visual distribution is independent of the text token. Then, the joint distribution factorizes as $p(v_i,t_j)=p(v_i)p(t_j)$ and $I(\mathbf{V};\mathbf{T})=0$. 
Consequently, the PAQ numerator vanishes for both modes, suppressing them to near-zero.

Only \textbf{optimal} heads exhibit query-conditioned sharp peaks where $\hat{p}(v_i \mid t_j)$ varies with $t_j$, concentrating probability mass on task-relevant visual tokens. This yields $I(\mathbf{V};\mathbf{T})>0$ and keeps $H(\mathbf{V})$ low by concentrating attention on a small token subset. The resulting high PAQ score uniquely distinguishes optimal heads with quantitative evidence in Appendix~\ref{app:mechanism}.

\subsection{HAP}
\label{sec:HAP}
Building upon the PAQ score (Eq.~\eqref{eq:PAQ}), we present HAP, a \textbf{h}ead-\textbf{a}daptive visual token \textbf{p}runing framework that requires no per-task hyper-parameter tuning.

HAP first partitions the $L$ transformer layers into $M$ consecutive groups. Each group has $L/M$ layers.
Prior works~\cite{xing2024pyramiddrop} observe that visual tokens naturally follow a pyramid structure, with token volume declining from lower to higher layers as features consolidate. 
HAP inherits this observation and formalizes it through a geometric decay: the target budget for group $g$ is $N_v(g) = V / (1+g)^2$, where $V$ is the initial number of visual tokens. Consequently, the stage count $M$ is derived directly from the target FLOPs budget via a closed-form cost function $\mathrm{F}(M)$ (Eq.~\ref{eq:flops_closed_form}), eliminating the need for per-task tuning.
We emphasize the distinction between \emph{hyper-parameters} (quantities requiring per-task tuning or grid search) and \emph{structural design choices}: the stage count $M$, group budgets $N_v(g)$, and PAQ-softmax weights are all derived deterministically from the single FLOPs-budget input, without any empirical search.

Then, HAP progressively aggregates optimal heads with prompt-grounded attention in groups through a PAQ-guided weighting scheme. 
Consider a candidate set $\mathcal{C}$, where each element $c \in \mathcal{C}$ represents cross-modal attention, and is associated with PAQ score $\mathrm{PAQ}_c$. 
To derive aggregation weights, we center the raw scores by subtracting the mean:
\begin{equation}
    s_c = \mathrm{PAQ}_c - \frac{1}{|\mathcal{C}|}\sum_{j\in\mathcal{C}}\mathrm{PAQ}_j,
    \label{eq:centering}
\end{equation}
where $s_c$ denotes the centered score for candidate $c$. 
We then apply standard softmax to obtain the aggregation weight:
\begin{equation}
    w_c = \frac{\exp(s_c)}{\sum_{j\in\mathcal{C}}\exp(s_j)}.
    \label{eq:softmax_weight}
\end{equation}
This yields a soft, self-normalized distribution. 
Candidates with higher PAQ naturally dominate, while suboptimal ones are down-weighted. 
The aggregated attention map is then computed as:
\begin{equation}
    \mathbf{A}_{\mathrm{agg}} = \sum_{c\in\mathcal{C}} w_c \cdot \mathbf{A}_c.
    \label{eq:unified_agg}
\end{equation}

Based on this, HAP first aggregates all heads within layer $l$. 
For layer $l$, we compute a PAQ score for each head from its attention map $\mathbf{A}_h^{(l)}$. 
Treating these $H$ heads as a candidate set, we obtain weights via Eq.~\eqref{eq:centering}--\eqref{eq:softmax_weight} and fuse their maps to produce the layer attention $\mathbf{A}_{\mathrm{agg}}^{(l)}$.
HAP then aggregates these layer attentions within each group. 
Next, we treat $L/M$ layers of each group as a new candidate set. Each candidate has an aggregated attention map $\mathbf{A}_{\mathrm{agg}}^{(l)}$ and a PAQ score. We apply the same PAQ-softmax procedure. Fusing these layer maps yields the final group attention $\mathbf{A}_{\mathrm{group}}^{(g)}$ for token scoring, effectively amplifying signals from optimal heads.

Unlike prior methods that select fixed anchor layers and average all heads uniformly, HAP softly aggregates optimal heads in groups. These cross-modal attentions ensure that visual tokens are scored by attention signals that more faithfully reflect prompt relevance. A comprehensive mechanism-level analysis of PAQ (weight non-uniformity, head-mode distributions, grounding correctness, and factorial ablations) is provided in Appendix~\ref{app:mechanism}.

Finally, we compute an importance score for each visual token by averaging over all text positions for group $g$:
\begin{equation}
    s_i^{(g)} = \frac{1}{N_t}\sum_{j=1}^{N_t} 
    \mathbf{A}_{\mathrm{group}}^{(g)}(j,i).
    \label{eq:token_score}
\end{equation}
 We sort visual tokens by $s_i^{(g)}$ and retain exactly $N_v(g)$ tokens with the highest scores throughout the prefill phase.

\subsection{Efficiency and Compatibility}
\label{sec:analysis}
Following PDrop~\cite{xing2024pyramiddrop}, the FLOPs of a transformer layer is:
\begin{equation}
    \mathrm{F}(N_v) \approx 4 N_v D^2 + 2 N_v^2 D + 3 N_v D d_{\mathrm{ffn}},
    \label{eq:layer_flops}
\end{equation}
where $N_v$, $D$, and $d_{\mathrm{ffn}}$ denote the visual token count, hidden dimension, and FFN intermediate dimension, respectively.

For PAQ computation and token pruning, we directly reuse the attention matrices from the forward pass. Concretely, computing the PAQ score for one head requires marginal aggregation over $N_t \times N_v(l)$ attention weights, incurring $\mathcal{O}(N_t N_v(l))$ operations; with $H$ heads per layer, the per-layer PAQ cost is $\mathcal{O}(H N_t N_v(l))$. Head aggregation via Eq.~\eqref{eq:centering}--\eqref{eq:softmax_weight} adds $\mathcal{O}(H)$ and is absorbed into this term. Token selection averages the aggregated attention over $N_t$ text positions and sorts $N_v(l)$ visual tokens, contributing $\mathcal{O}(N_t N_v(l) + N_v(l)\log N_v(l))$. Layer aggregation within each group repeats the PAQ-softmax over $L/m$ layers, adding $\mathcal{O}((L/m) N_t N_v(l))$ per group. Because $N_v(l)$ decays geometrically with depth, summing these components across all $L$ layers bounds the total overhead by $\mathcal{O}(H N_t V L/m)$. This remains strictly smaller than the unpruned attention cost and, under typical configurations where $D \gg H$ and $V \gg N_t$, constitutes a lower-order term relative to the full forward-pass FLOPs. Specifically, we extend the dual-flash strategy from SparseVLM~\cite{zhang2024sparsevlm} to compute per-head cross-modal distributions without materializing the full $\mathcal{O}(N^2)$ attention map (see Appendix~\ref{app:flash}). Taken together with the pruned transformer costs, the total computational cost is fully characterized by the following closed-form function of the stage count $m$ (see Appendix~\ref{app:derivation} for the full derivation):
\begin{equation}
\begin{split}
    \mathrm{F}_{\mathrm{pruned}}(m) \approx \frac{L}{m}\left[ 
    A\!\left(1 - \frac{1}{1+m}\right) \right. \\
    \left. + \frac{B}{3}\!\left(1 - \frac{1}{(1+m)^3}\right) \right],
\end{split}
\label{eq:flops_closed_form}
\end{equation}
where $A = 4VD^2 + 3VDd_{\mathrm{ffn}}$ and $B = 2V^2D$ are architecture constants. Equation~\eqref{eq:flops_closed_form} confirms that the cost is a deterministic function of $m$ alone; no architecture-specific pruning schedule or per-task calibration is required.

\section{Experiments}
\subsection{Experimental Settings}
\label{exp}

To assess the effectiveness of our method, we conduct experiments on LLaVA-1.5-7B, Qwen2.5-VL-7B, Qwen3-VL-8B~\cite{bai2025qwen3}, InternVL3-8B and DeepSeek-VL2 Small-16B across 18 benchmarks. These include grounding and OCR tasks:  TextVQA$^{\text{val}}$~\cite{singh2019towards},  RefCOCO, RefCOCO+~\cite{kazemzadeh2014referitgame}, and RefCOCOg~\cite{mao2016generation}; comprehensive VQA benchmarks: Seed2-Plus~\cite{li2024seed}, AI2D~\cite{kembhavi2016diagram}, MMStar~\cite{chen2024we}, RealWorldQA~\cite{grok15v}, SQA$^{\text{IMG}}$~\cite{lu2022learn}, VizWiz~\cite{chen2022grounding}, MMVet~\cite{yu2023mm}, POPE~\cite{li2023evaluating}, MME, MMBench$^{\text{EN/CN}}$ (MMB$^{\text{EN/CN}}$)~\cite{liu2024mmbench}, and GQA~\cite{hudson2019gqa}; and video benchmarks: MVBench~\cite{li2024mvbench} and MLVU~\cite{zhou2024mlvu}. More details of benchmarks are in Appendix~\ref{app:dataset}. We compare HAP with cross-modal attention methods (e.g., FastV~\cite{chen2024image}, SparseVLM~\cite{zhang2024sparsevlm}, PDrop~\cite{xing2024pyramiddrop}, AutoPrune~\cite{wang2026each}) and visual attention methods (e.g., VisionZip~\cite{yang2025visionzip}, VisPruner~\cite{zhang2025beyond}, CDPruner~\cite{zhang2026beyond}, IVC-Prune~\cite{sun2026ivc}). For fair comparison, we unify the token budget and compare FLOPs, KV cache, and inference speed. All experiments are conducted on a single NVIDIA RTX~3090. Unless otherwise stated, results are reported from a single run; statistical robustness over 10 repeated runs is analyzed in Appendix~\ref{app:statistics}.
\begin{table*}[t]
\caption{Evaluation of our method on the LLaVA-1.5-7B model across nine datasets under three visual token compression levels (192, 128, and 64). The vanilla configuration uses 576 tokens and average 3.82T FLOPs. Relative score is the average ratio between the score and original score across all benchmarks.}
\label{tab:llava}
\centering
\vspace{-3mm}
\resizebox{\textwidth}{!}{
\begin{tabular}{@{}lccccccccccccc@{}}
\toprule
\textbf{Method} & \textbf{Venue} & \textbf{SQA} & \textbf{GQA} & \textbf{MMB$^{\text{EN}}$} & \textbf{MMB$^{\text{CN}}$} & \textbf{POPE} & \textbf{Text} & \textbf{MME} & \textbf{VizWiz} & \textbf{MMVet} & \textbf{\makecell{Rel.\\Score}} & \textbf{FLOPs} \\
\midrule
Original & - & 69.5 & 61.9 & 64.6 & 58.1 & 86.1 & 58.2 & 1864 & 50.0 & 30.9 & \textbf{100.0\%} & 3.82T \\
\midrule
\multicolumn{13}{@{}l@{}}{\textit{\textbf{Retain 128 Tokens (22\% Retention)}}} \\
FastV & ECCV'24 & 68.6 & 49.6 & 56.1 & 56.4 & 53.4 & 50.5 & 1490 & 51.3 & 26.3 & 86.6\% & 1.64T \\
SparseVLM & ICML'25 & 67.1 & 56.0 & 60.0 & 51.1 & 80.5 & 54.9 & 1696 & 51.4 & 29.6 & 93.9\% & 1.66T \\
PDrop & CVPR'25 & \textbf{69.9} & 56.0 & 61.1 & 56.6 & 82.3 & 55.1 & 1664 & 51.0 & 30.8 & 96.0\% & 1.53T \\
VisionZip & CVPR'25 & 68.9 & 57.6 & 62.0 & 53.4 & 83.2 & 56.8 & 1763 & 51.2 & 30.2 & 96.6\% & 1.44T \\
VisPruner & ICCV'25 & 68.2 & 57.4 & 62.7 & 57.3 & 84.6 & 57.0 & 1789 & 51.7 & 30.7 & 97.9\% & 1.45T \\
AutoPrune & ICCV'25 & 69.2 & 59.5 & 64.3 & 57.0 & 84.5 & 57.1 & 1785 & 49.1 & 30.3 & 98.0\% & 1.48T \\
\rowcolor{cyan!8}
\textbf{HAP} & - & 69.7 & \textbf{62.1} & \textbf{64.6} & \textbf{59.0} & \textbf{86.4} & \textbf{58.5} & \textbf{1872} & \textbf{51.8} & \textbf{33.4} & \textbf{102\%} & 1.42T \\
\midrule
\multicolumn{13}{@{}l@{}}{\textit{\textbf{Retain 64 Tokens (11\% Retention)}}} \\
FastV & ECCV'24 & 68.7 & 46.1 & 47.2 & 52.7 & 38.2 & 47.8 & 1255 & 50.8 & 19.6 & 77.3\% & 1.24T \\
SparseVLM & ICML'25 & 62.2 & 52.7 & 56.2 & 46.1 & 75.1 & 51.8 & 1505 & 50.1 & 23.3 & 86.0\% & 1.24T \\
PDrop & CVPR'25 & \textbf{69.2} & 43.9 & 53.3 & 50.5 & 75.9 & 48.9 & 1561 & 50.7 & \textbf{30.7} & 90.4\% & 1.17T \\
VisionZip & CVPR'25 & 69.0 & 55.1 & 60.1 & 51.9 & 77.0 & 55.5 & 1690 & 51.2 & 29.6 & 93.8\% & 1.12T \\
VisPruner & ICCV'25 & 69.1 & 55.4 & 61.3 & 55.1 & 80.4 & 55.8 & 1706 & 53.3 & 32.3 & 96.7\% & 1.14T \\
AutoPrune & ICCV'25 & 68.9 & 57.1 & 63.4 & 56.2 & 83.3 & 56.9 & 1745 & 49.4 & 29.9 & 96.9\% & 1.15T \\
\rowcolor{cyan!8}
\textbf{HAP} & - & 69.3 & \textbf{60.3} & \textbf{64.5} & \textbf{59.6} & \textbf{86.0} & \textbf{58.3} & \textbf{1883} & \textbf{52.1} & 30.2 & \textbf{100.3\%} & 1.12T \\
\midrule
\multicolumn{13}{@{}l@{}}{\textit{\textbf{Retain 32 Tokens (5.6\% Retention)}}} \\
FastV & ECCV'24 & 42.6 & 41.5 & 37.8 & 33.2 & 32.5 & 42.5 & 1184 & 51.7 & 20.7 & 65.4\% & 1.02T \\
SparseVLM & ICML'25 & 57.3 & 48.3 & 51.4 & 40.6 & 67.9 & 46.1 & 1347 & 51.9 & 18.6 & 78.3\% & 1.10T \\
PDrop & CVPR'25 & 67.5 & 40.3 & 51.4 & 48.9 & 68.6 & 45.7 & 1492 & 48.5 & 25.1 & 82.6\% & 0.98T \\
VisionZip & CVPR'25 & 67.2 & 51.5 & 57.3 & 49.1 & 74.3 & 50.9 & 1486 & 48.7 & 26.8 & 93.8\% & 0.91T \\
VisPruner & ICCV'25 & 69.2 & 52.2 & 58.4 & 52.7 & 72.7 & 53.9 & 1691 & \textbf{53.0} & 28.8 & 92.4\% & 0.93T \\
AutoPrune & ICCV'25 & 65.7 & 58.5 & 61.0 & 54.9 & 81.4 & 55.0 & 1732 & 49.7 & \textbf{29.2} & 94.9\% & 1.04T \\
\rowcolor{cyan!8}
\textbf{HAP} & - & \textbf{69.2} & \textbf{60.0} & \textbf{64.3} & \textbf{58.9} & \textbf{85.5} & \textbf{57.5} & \textbf{1875} & 51.7 & 28.6 & \textbf{99.1\%} & 0.89T \\
\bottomrule
\end{tabular}
}
\vspace{-4mm}
\end{table*}

\begin{table*}[h]
\begin{center}
\caption{Performance comparison on InternVL3 and Qwen2.5-VL VLMs with 25\% visual tokens retained.}
\resizebox{\textwidth}{!}{
\begin{tabular}{clcccccccccc}
\hline
\textbf{Model} & \textbf{Method} & \textbf{Venue} & \textbf{MMB\textsuperscript{EN}} & \textbf{MME} & \textbf{POPE} & \textbf{MVBench} & \textbf{AI2D} & \textbf{Seed2} & \textbf{RWQA} & \textbf{\makecell{Relative\\Score(\%)}} & \textbf{FLOPs(T)} \\
\hline
\multirow{5}{*}{\centering InternVL3-8B} 
& original & - & 83.4 & 2415 & 91.1 & 75.4 & 85.2 & 69.7 & 71.6 & 100\% & 8.15 \\
& FastV & ECCV'24 & 80.4 & 2235 & 87.6 & 72.9 & 82.2 & 62.5 & 62.6 & 91.5\% (↓8.5\%) & 3.05 \\
& PDrop & CVPR'25 & 82.1 & 2298 & 88.2 & 73.4 & 83.1 & 63.8 & 63.8 & 94.2\% (↓5.8\%) & 3.02 \\
& VisPruner & ICCV'25 & 81.3 & 2372 & 89.4 & 72.1 & 82.2 & 64.9 & 66.1 & 95.3\% (↓4.7\%) & 2.98 \\
& \cellcolor{cyan!8}\textbf{HAP} & \cellcolor{cyan!8}- & \cellcolor{cyan!8}\textbf{83.8} & \cellcolor{cyan!8}\textbf{2417} & \cellcolor{cyan!8}\textbf{90.6} & \cellcolor{cyan!8}\textbf{74.2} & \cellcolor{cyan!8}\textbf{83.7} & \cellcolor{cyan!8}\textbf{69.1} & \cellcolor{cyan!8}\textbf{71.9} & \cellcolor{cyan!8}\textbf{99.5\% (\textcolor{red}{↓0.5\%})} & \cellcolor{cyan!8}2.90 \\
\hline
\hline
\textbf{Model} & \textbf{Method} & \textbf{Venue} & \textbf{MMB\textsuperscript{EN}} & \textbf{MME} & \textbf{POPE} & \textbf{MMB\textsuperscript{CN}} & \textbf{AI2D} & \textbf{MLVU} & \textbf{RWQA} & \textbf{\makecell{Relative\\Score(\%)}} & \textbf{FLOPs(T)} \\
\hline
\multirow{4}{*}{\centering Qwen2.5-VL-7B} 
& original & - & 83.5 & 2305 & 86.2 & 83.4 & 81.1 & 75.0 & 68.1 & 100\% & 6.10 \\
& CDPruner & NeurIPS'25 & 64.2 & 2200 & 72.5 & 79.0 & 56.3 & 63.1 & 57.3 & 84.1\% (↓15.9\%) & 2.60 \\
& IVC-Prune & ICLR'26 & 79.1 & 2183 & 81.6 & 79.0 & 76.8 & 71.0 & 64.5 & 94.7\% (↓5.3\%) & 2.58 \\
& \cellcolor{cyan!8}\textbf{HAP} & \cellcolor{cyan!8}- & \cellcolor{cyan!8}\textbf{83.3} & \cellcolor{cyan!8}\textbf{2293} & \cellcolor{cyan!8}\textbf{85.9} & \cellcolor{cyan!8}\textbf{82.7} & \cellcolor{cyan!8}\textbf{80.3} & \cellcolor{cyan!8}\textbf{74.3} & \cellcolor{cyan!8}\textbf{68.6} & \cellcolor{cyan!8}\textbf{99.1\% (\textcolor{red}{↓0.9\%})} & \cellcolor{cyan!8}2.55 \\
\hline
\end{tabular}
}
\label{tab:intern_qwen}
\end{center}
\vspace{-5mm}
\end{table*}

\begin{table*}[t]
\caption{\textbf{Performance comparisons on Qwen-3-VL-8B} across 3 grounding benchmarks under 75\% pruning ratio.}
\centering
\resizebox{\textwidth}{!}{
\begin{tabular}{@{}ll|ccc|ccc|cc|c@{}}
\toprule
\multirow{2}{*}{\textbf{Method}} & \multirow{2}{*}{\textbf{Venue}} & \multicolumn{3}{c|}{\textbf{RefCOCO}} & \multicolumn{3}{c|}{\textbf{RefCOCO+}} & \multicolumn{2}{c|}{\textbf{RefCOCOg}} & \multirow{2}{*}{\textbf{Average}} \\
\cmidrule(lr){3-5} \cmidrule(lr){6-8} \cmidrule(lr){9-10}
& & val & testA & testB & val & testA & testB & val & test & \\
\midrule
\rowcolor{gray!15}
Qwen3-VL-8B  & - & 91.58 & 93.27 & 86.76 & 85.80 & 90.22 & 79.78 & 88.76 & 88.24 & 100.0\% \\
\midrule
FastV  & ECCV'24 & 43.57 & 46.81 & 40.86 & 39.47 & 43.78 & 36.02 & 43.04 & 42.69 & 47.7\% \\
PDrop & CVPR'25 & 46.46 & 53.83 & 37.23 & 42.29 & 47.76 & 32.81 & 45.32 & 44.91 & 49.8\% \\
\rowcolor{cyan!8}
\textbf{HAP} & - & \textbf{76.02} & \textbf{80.65} & \textbf{70.02} & \textbf{68.72} & \textbf{75.62} & \textbf{60.65} & \textbf{70.82} & \textbf{71.23} & \textbf{81.5\%} \\
\bottomrule
\end{tabular}
}
\label{tab:refcoco_comparison}
\vspace{-2mm}
\end{table*}

\subsection{Main Results}
In Tab.~\ref{tab:llava}, we evaluate HAP on LLaVA-1.5-7B across nine image understanding tasks. When reducing visual tokens from 576 to 32, HAP incurs only a 0.9\% overall performance drop and 0.89 TFLOPs, substantially lower than competitors. Methods relying directly on cross-modal attention (\emph{e.g.}, FastV, SparseVLM, and PDrop) suffer degradation exceeding 15\%. Unimodal approaches such as VisionZip and VisPruner, lacking textual guidance, lose task-relevant regions at extreme compression ratios, incurring over 6\% performance drop. AutoPrune employs mutual information for layer-wise budget allocation, neglecting the compression term essential for assessing cross-modal alignment quality and thus failing to identify reliable pruning signals with 5\% performance drop. Notably, HAP improves results on MME, VizWiz, and MMB$^{\mathrm{CN}}$ when retaining 32 tokens, indicating accurate removal of hallucination-inducing visual tokens. A systematic retention-ratio sweep with mean and standard deviation over 10 runs (Appendix~\ref{app:statistics}) further confirms that these gains are stable and not attributable to run-to-run variance.

In Tab.~\ref{tab:intern_qwen}, HAP achieves 99.5\% and 99.1\% relative scores on InternVL3-8B and Qwen2.5-VL-7B at 25\% token retention, outperforming FastV, PDrop, and VisPruner (all $>$4\% drops). On the challenging REC task (Tab.~\ref{tab:refcoco_comparison}), HAP attains \textbf{81.5\%} on Qwen3-VL-8B under 75\% pruning, surpassing FastV and PDrop by over 30\% absolute margins. Tab.~\ref{tab:efficiency_internvl} further shows HAP delivers a \textbf{1.96$\times$} speedup on InternVL3-8B with only 34\% KV cache overhead and 99.7\% accuracy. Tab.~\ref{tab:deepseek_HAP} in  Appendix~\ref{app:deepseek} further reports that HAP attains a 103.0\% relative average on DeepSeek-VL2 Small-16B~\cite{wu2024deepseek} with 50\% token retention, outperforming the baselines across all six general VQA benchmarks.

\begin{table}[t]
\centering
\caption{Efficiency with InternVL3-8B at 25\% tokens.}
\label{tab:efficiency_internvl}
\setlength{\tabcolsep}{1pt}
\begin{tabular}{@{}l@{\hspace{-6pt}}ccc@{}}
\toprule
\textbf{Method} & \textbf{KV Cache} & \textbf{Perf.} & \textbf{Speedup} \\
& \textbf{(\%)} & \textbf{(\%)} & \textbf{($\times$)} \\
\midrule
FastV & 52 & 92.5 & 1.54 \\
VisPruner & 37 & 95.8 & 1.87 \\
HAP (Ours) & \textbf{34} & \textbf{99.7} & \textbf{1.96} \\
\bottomrule
\end{tabular}
\end{table}

\subsection{Ablation Study}

\begin{figure}[t]
\centering
\includegraphics[width=\linewidth]{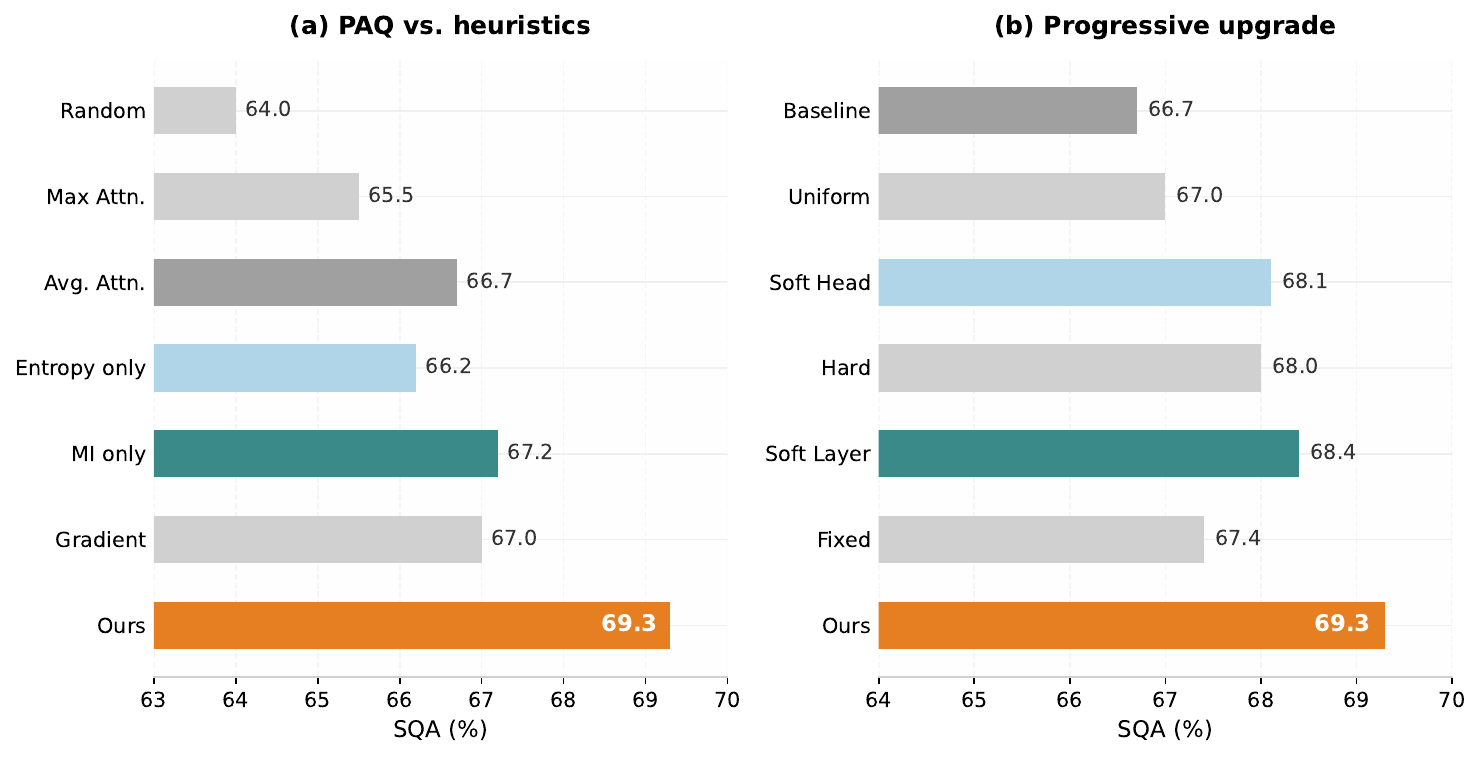}
\vspace{-2mm}
\includegraphics[width=\linewidth]{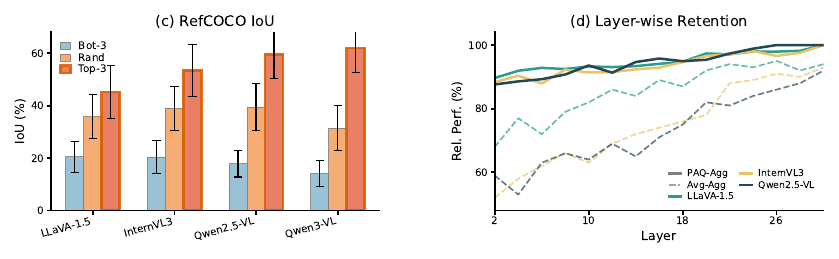}
\caption{(a)~PAQ \emph{vs.}~heuristic criteria on SQA. (b)~Progressive integration of HAP components. (c)~RefCOCO IoU of tokens selected by PAQ ranking across four VLMs. (d)~Layer-wise retention stability (25\% tokens): PAQ aggregation (solid) maintains robust performance across depths, while uniform aggregation (dashed) suffers from severe early-layer degradation caused by sink/dispersed heads.}
\label{fig:ablation}
\vspace{-5mm}
\end{figure}

\paragraph{Ablation of HAP.}
Fig.~\ref{fig:ablation} shows that the gains stem from PAQ's joint formulation rather than heuristics.
(a)~PAQ (69.3\%) substantially outperforms random pruning (64.0\%), max/average attention (65.5\% / 66.7\%), and isolated criteria such as entropy-only (66.2\%) and MI-only (67.2\%). Gradient-based saliency (67.0\%, computed via backpropagated gradients) further verifies that PAQ achieves superior alignment without backpropagation.
(b)~Progressive integration yields monotonic improvements: soft head aggregation raises the baseline to 68.1\%, soft layer fusion reaches 68.4\%, and the full framework attains 69.3\%. Replacing soft PAQ-weighted aggregation with uniform weighting, hard top-$k$ selection, or a fixed per-group budget degrades results, confirming the principled design. A strict factorial ablation that isolates PAQ weighting from the geometric budget schedule is reported in Appendix~\ref{app:mechanism}.
(c)~RefCOCO IoU validates that PAQ identifies spatially critical tokens: top-PAQ candidates consistently achieve higher IoU than random or bottom-PAQ baselines across all four architectures, with the gap widening on stronger models.
(d)~PAQ mitigates layer-wise heterogeneity: when retaining only 25\% of visual tokens at each layer independently, PAQ-guided aggregation (solid curves) maintains stable relative performance ($>$85\%) across all depths, whereas uniform aggregation (dashed curves) exhibits severe fluctuations---particularly in early layers where sink and dispersed heads dominate (e.g., LLaVA-1.5 drops below 55\% at Layer 4). This confirms that PAQ reliably selects optimal attention patterns regardless of layer depth, eliminating the performance volatility that plagues fixed-layer pruning strategies (Fig.~\ref{fig1}a).

\begin{figure*}[ht]
\centering
\includegraphics[width=1.0\linewidth]{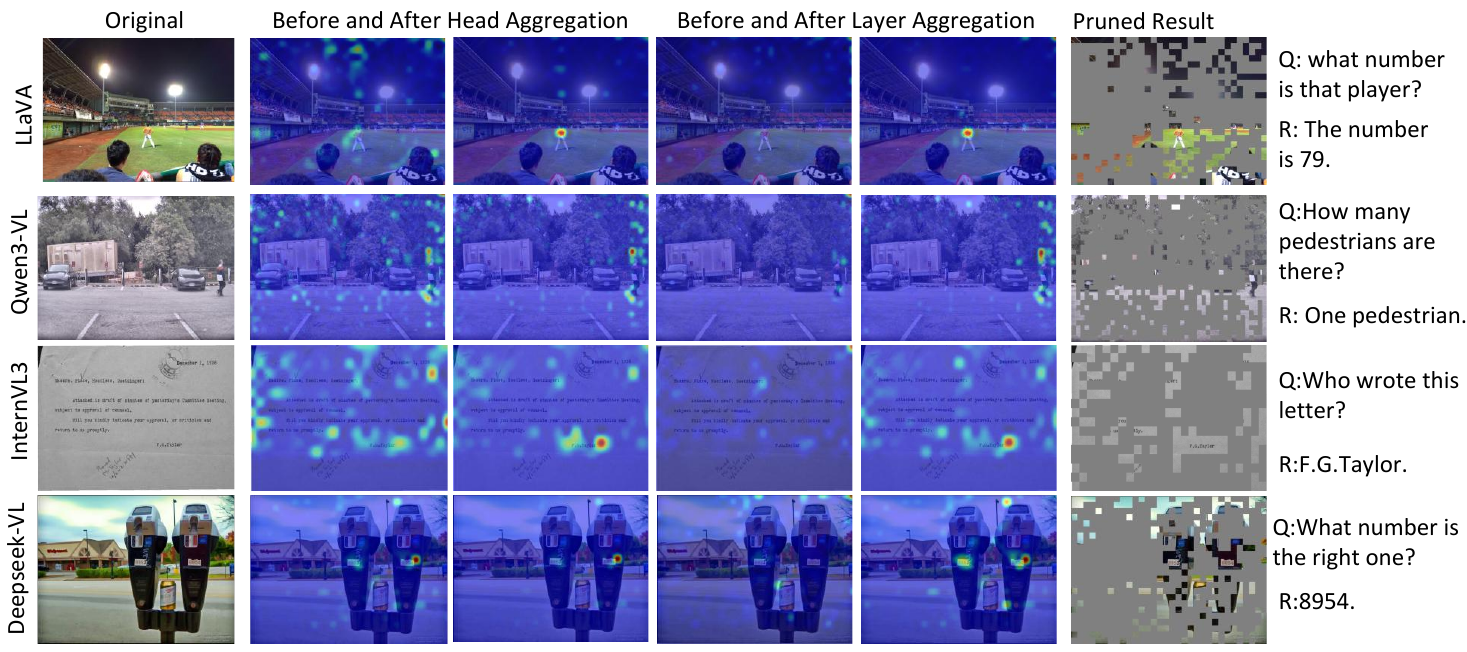}
\caption{Qualitative visualization of HAP across LLaVA, Qwen3-VL, InternVL3, and DeepSeek-VL. Columns from left: original image; attention maps before/after PAQ-weighted head aggregation and before/after PAQ-weighted layer aggregation; final pruned token masks; and corresponding QA pairs.}
\label{fig:visualization}
\vspace{-3mm}
\end{figure*}
Additionally, we evaluate HAP's robustness to prompt variations. We test three prompt styles: \textit{standard} (original benchmark prompt), \textit{detailed} (augmented with task-relevant image descriptions, e.g., scene context and object-relationship hints to encourage visual grounding), and \textit{corrupt} (adversarially perturbed by random word substitution and key-token deletion to inject semantic noise). As shown in Table~\ref{tab:prompt_robustness}, all scores are reported as relative performance (\%) against the original prompt. Despite the detailed prompt providing helpful visual context and the corrupt prompt introducing deliberate lexical interference, the maximum variation stays within 1.2\%, confirming that HAP's pruning decisions are driven by intrinsic cross-modal alignment rather than superficial prompt artifacts. We further evaluate HAP under systematic image corruptions and multi-turn dialogues in Appendix~\ref{app:promptdiverse}.

\begin{table}[h]
\centering
\small
\setlength{\tabcolsep}{6pt}
\caption{Prompt robustness of HAP (LLaVA-1.5-7B, 128 tokens retained).
All scores are relative performance (\%) against the original prompt.
$\Delta$ denotes the maximum relative variation across prompt styles.}
\vspace{-2mm}
\label{tab:prompt_robustness}
\begin{tabular}{lccc|c}
\toprule
\textbf{Benchmark} & \textbf{Standard} & \textbf{Detailed} & \textbf{Corrupt} & $\mathbf{\Delta}$ (\%) \\
\midrule
SQA     & 100.3 & 100.7 &  99.3 & 1.0 \\
GQA     & 100.3 & 100.8 &  99.2 & 1.1 \\
POPE    & 100.3 & 100.6 &  99.4 & 0.9 \\
MME     & 100.4 & 100.9 &  99.3 & 1.2 \\
\midrule
\textbf{Average} & \textbf{100.3} & \textbf{100.8} & \textbf{99.3} & \textbf{1.0} \\
\bottomrule
\end{tabular}
\vspace{-3mm}
\end{table}

\section{Visualization Analysis}
\label{sec:appendix}

Fig.~\ref{fig:visualization} validates HAP on four VLMs across scene reasoning, counting, OCR, and digit recognition. After PAQ-weighted head aggregation, attention maps sharpen from noisy mixtures to concentrated regions (e.g., jersey numbers and pedestrian silhouettes), confirming that optimal heads dominate while sink and dispersed patterns are suppressed. Layer aggregation further refines these maps into compact, semantically coherent regions such as signature blocks and digit strings. The final pruned masks retain only task-critical tokens, enabling correct answers despite aggressive compression and verifying that HAP's two-level aggregation generalizes robustly across architectures.
\paragraph{Efficiency Analysis.}In Tab.~\ref{tab:token_reduction}, we evaluate the practical efficiency of HAP on a single NVIDIA RTX 3090 (24GB). As visual tokens are progressively compressed, both latency and KV cache usage decrease substantially. Reducing tokens from 576 to 128 decreases latency from 0.48~s to 0.22~s, yielding a \textbf{2.18$\times$} speedup while KV cache drops to 36\%. Further compression to 64 tokens achieves \textbf{2.53$\times$} speedup (0.19~s) with only 25\% KV cache usage, retaining 99.7\% performance. At 32 tokens, latency reduces to 0.17~s (\textbf{2.82$\times$} speedup) with 20\% KV cache, maintaining 99.1\% accuracy. These results demonstrate that HAP effectively balances computational efficiency and model fidelity, substantially reducing memory and runtime demands with minimal performance degradation. A breakdown of prefill versus decoding latency, batch-size scaling, and long-generation regimes is provided in Appendix~\ref{app:decode}.

\section{Conclusion}
\label{sec:conclusion}
\vspace{-2mm}
\begin{table}[t]
\centering
\caption{Performance, latency, and KV cache usage analysis.}
\label{tab:token_reduction}
\vspace{-2mm}
\setlength{\tabcolsep}{5pt}
\begin{tabular}{lcccc}
\toprule
Retained tokens & 576 & 128 & 64 & 32 \\ \midrule
Performance (\%) & 100 & 102 & 99.7 & 99.1 \\
KV Cache (\%) & 100 & 36 & 25 & 20 \\
Latency (s) & 0.48 & 0.22 & 0.19 & 0.17 \\
Speedup ($\times$) & 1.00 & 2.18 & 2.53 & 2.82 \\
\bottomrule
\end{tabular}
\vspace{-3mm}
\end{table}

This paper presents HAP, a training free framework for visual token pruning in Vision Language Models. We identify heterogeneous cross-modal alignment across heads and layers, and propose the Prompt Grounded Attention Quality (PAQ) score to unify reliability assessment at both levels. HAP dynamically fuses attention heads and layer consensus via PAQ guided soft aggregation, allocating pruning budgets by alignment quality. Experiments across five architectures and 18 benchmarks demonstrate state of the art performance efficiency trade offs without hyperparameter tuning.

\section*{Acknowledgements}
This work is funded by NSFC (No. 62525209, T2421002, 623B2071), National Key
Laboratory of Data Space Technology
and System and ZTE Industry-University-Research Pre-research Fund.

\section{Limitations}
HAP leaves several natural directions for extension. First, our framework operates during prefill and leaves the visual KV cache untouched during decoding. Extending token consolidation to the autoregressive stage could further reduce memory for long outputs. A quantitative analysis of this boundary (Appendix~\ref{app:decode}) shows that FLOPs savings remain positive up to decode lengths on the order of $10^{5}$ tokens, and that the reduced visual KV cache continues to benefit decoding throughput beyond that point. Second, PAQ is an attention-based proxy derived from cross-modal attention, which assumes paired text queries. Although we provide correlational evidence linking high PAQ to correct visual grounding (Appendix~\ref{app:mechanism}), fully causal validation (e.g., counterfactual token interventions) and applying similar head-adaptive criteria to unimodal compression remain future work. Finally, our evaluation focuses on static-image and short-video benchmarks. Head-level alignment patterns in streaming or ultra-long multimodal contexts warrant additional study.

Additionally, we acknowledge the use of AI-assisted writing tools (e.g., Claude) for language polishing, grammar correction, and stylistic refinement during manuscript preparation. These tools were not used to generate core technical contents such as research ideas, experimental results.
\bibliography{custom}
\appendix

\section{Cross-Modal Head Patterns versus LLM Attention Sinks}
\label{app:llm_vlm}

Attention head heterogeneity has been extensively studied in text-only LLMs. Attention sinks concentrate probability mass on delimiter or initial tokens~\cite{xiao2023efficient}, and surveys of head-level functional specialization document a variety of syntactic, positional, and retrieval-oriented head roles~\cite{zheng2024attention}. Cross-modal attention in VLMs exhibits qualitatively distinct characteristics. LLM sinks anchor on textual delimiter tokens, whereas VLM visual sinks fixate on image backgrounds or edges~\cite{kang2025see}; these are fundamentally different physical meanings, so LLM-side theories of sink preservation do not transfer directly to cross-modal pruning. Moreover, VLM heads follow heterogeneous routing behaviors, alternating between ``visual sinks'' that aggregate visual tokens and dispersed heads that distribute attention broadly~\cite{luo2025tosink}.

This structural difference motivates HAP's design. Whereas LLM-oriented methods typically preserve sink tokens to stabilize generation, PAQ actively \emph{down-weights} misaligned heads, ensuring that only prompt-relevant heads dominate the pruning signal. The head-mode taxonomy in Fig.~\ref{fig0} (sink, dispersed, optimal) is thus a cross-modal phenomenon with no direct counterpart in unimodal self-attention, and the quantitative correspondence between PAQ and these modes is established in Appendix~\ref{app:mechanism:correspondence}.

\section{Mechanism Analysis of PAQ}
\label{app:mechanism}

This appendix provides the mechanism-level evidence behind PAQ, complementing the downstream results in the main paper. All measurements are conducted on LLaVA-1.5-7B and Qwen2.5-VL-7B.

\subsection{Non-uniformity of PAQ-Softmax Weights}
\label{app:mechanism:nonuniform}

A natural concern is that, since PAQ is bounded in $[0,1]$, head-level PAQ scores within a layer might concentrate in a narrow range, making the centered softmax weights (Eq.~\eqref{eq:softmax_weight}) close to the uniform weight $1/H$. Our measurements refute this. Across all layers of LLaVA-1.5-7B and Qwen2.5-VL-7B, the mean intra-layer standard deviation of PAQ is 0.28 and 0.31 against mean PAQ values of 0.64 and 0.59, corresponding to coefficients of variation of approximately 44\% and 53\%. Consequently, the effective number of heads, $N_{\mathrm{eff}} = 1/\sum_h w_h^2$, averages only 4.2 and 5.1, far below the total head count. The divergence from uniform weighting, measured by $\mathrm{KL}(w \,\|\, \mathrm{uniform})$, averages 0.83 nats on LLaVA-1.5-7B and 1.12 nats on Qwen2.5-VL-7B, with maxima exceeding 2.0 nats in early layers where sink and dispersed heads are abundant. These results confirm that PAQ-softmax produces strongly non-uniform, concentrated aggregation weights rather than a mild perturbation of uniform averaging.

\subsection{Correctness of High-PAQ Heads}
\label{app:mechanism:correctness}

To test whether high PAQ reflects correct visual grounding rather than spurious prompt sensitivity, we compute the per-head correlation between PAQ and RefCOCO IoU. The Pearson correlation is 0.67 on LLaVA-1.5-7B and 0.71 on Qwen2.5-VL-7B ($p < 0.001$). Pruning with only the top-30\% PAQ heads attains RefCOCO IoU of 78.4\% and 81.2\% on the two models, whereas using only the bottom-30\% heads drops IoU to 52.1\% and 55.6\%. This is consistent with Fig.~\ref{fig:ablation}(c), where top-PAQ tokens consistently outperform random and bottom-PAQ baselines across all four evaluated VLMs.

We also inspected failure cases. Among the rare high-PAQ heads with low IoU ($<0.3$), 82\% attend to semantically related but non-target regions (\eg adjacent text or background objects of the same category) rather than random distractors. This indicates that high PAQ signals genuine prompt-relevant alignment, not arbitrary high responsiveness.

\subsection{Layer-Wise Head Modes and Budget Allocation}
\label{app:mechanism:layerwise}

We quantified the layer-wise proportion of sink, dispersed, and optimal heads. High-PAQ (optimal) heads are non-uniformly distributed across depth: they emerge predominantly in middle-to-deep layers (layers 8--24 in LLaVA-1.5-7B and 10--28 in Qwen2.5-VL-7B), while early layers (1--6) are dominated by sink and dispersed patterns. This depth-wise heterogeneity is consistent with the geometric pyramid schedule: early layers retain more tokens because fewer optimal heads are available to guide reliable pruning, whereas deeper layers possess strong alignment signals and tolerate aggressive reduction.

We further compared against a PAQ-adaptive budget schedule that allocates tokens proportionally to the mean PAQ of each group. It attains a 99.3\% relative score versus 99.1\% for the fixed geometric pyramid, but requires 0.08T extra FLOPs (11\% overhead) because non-uniform token counts disrupt the regular pyramid structure. The geometric schedule is therefore the superior Pareto choice.

\subsection{Factorial Ablation of HAP Components}
\label{app:mechanism:factorial}

To isolate the contribution of PAQ weighting from the budget schedule, we perform a strict factorial ablation on LLaVA-1.5-7B with 32 tokens retained (5.6\%), crossing the scoring rule (PAQ vs.\ uniform) with the budget schedule (geometric pyramid vs.\ uniform budget). Table~\ref{tab:factorial} reports the results.

\begin{table}[h]
\centering
\small
\caption{Factorial ablation on LLaVA-1.5-7B (32 tokens retained, 5.6\%). Scoring rule and budget schedule are varied independently; relative score (\%) is reported.}
\label{tab:factorial}
\setlength{\tabcolsep}{8pt}
\begin{tabular}{lcc}
\toprule
\textbf{Scoring} & \textbf{Budget} & \textbf{Rel. Score (\%)} \\
\midrule
PAQ & Pyramid (HAP) & \textbf{99.1} \\
Uniform & Pyramid & 89.4 \\
PAQ & Uniform & 96.8 \\
Uniform & Uniform & 86.2 \\
\bottomrule
\end{tabular}
\vspace{-3mm}
\end{table}

The clear ranking $\text{PAQ+Pyramid} > \text{PAQ+Uniform} > \text{Uniform+Pyramid} > \text{Uniform+Uniform}$ shows that PAQ weighting provides the largest single gain ($\sim$9.7 points over uniform scoring under the same pyramid budget), while the geometric schedule adds a complementary $\sim$2.3 points. PAQ is thus the primary driver of performance, with the budget schedule acting as a secondary but consistent contributor.

\subsection{PAQ-to-Pattern Correspondence}
\label{app:mechanism:correspondence}

Finally, we quantified attention-map statistics by PAQ quintile on both models. In the top PAQ quintile, the average spatial entropy of the attention maps is 2.1 (LLaVA-1.5-7B) and 1.9 (Qwen2.5-VL-7B), with mean RefCOCO IoU of 74.3\% and 77.1\%. In the bottom quintile, spatial entropy rises to 5.8 and 6.2, while IoU drops to 31.2\% and 33.5\%. The relationship is monotonic across all quintiles (Spearman $\rho = -0.91$ and $-0.93$ between PAQ and spatial entropy). This confirms that PAQ directly indexes spatial concentration and semantic alignment quality---the sink/dispersed/optimal taxonomy of Fig.~\ref{fig0}---rather than merely prompt sensitivity.

\section{Compatibility with FlashAttention}
\label{app:flash}

Modern Vision-Language Models predominantly adopt FlashAttention~\cite{dao2022flashattention} or its variants as the default attention backend to eliminate the $O(N^2)$ memory bottleneck of standard self-attention. However, conventional token pruning methods that rely on per-token attention importance typically require materializing the full $|\mathcal{T}| \times |\mathcal{V}|$ cross-modal attention matrix, thereby forfeiting the memory benefits of FlashAttention. HAP circumvents this limitation by design: the PAQ scores (Eq.~\ref{eq:PAQ_full}) depend solely on \emph{aggregated} cross-modal statistics per layer, rather than on individual token-level attention weights.

\paragraph{Why full attention materialization is unnecessary.}
For each layer, PAQ reduces to a low-dimensional quantity that collapses the vision-token dimension. Concretely, let $\mathcal{T}$ and $\mathcal{V}$ denote the text and vision token sets. The required statistic takes the form of a reduction over vision tokens,
\[
\alpha_{\ell} \;=\; \sum_{j\in\mathcal{V}} f\!\left( \sum_{i\in\mathcal{T}} \mathrm{softmax}_j\!\left(\frac{Q_i K_j^\top}{\sqrt{d_k}}\right) \right),
\]
where $f(\cdot)$ is an aggregation function (e.g., $\max$, mean, or log-sum-exp) dictated by the PAQ formulation. Because the inner softmax normalizes over text queries and the outer sum collapses the vision keys, the result is a \emph{scalar per layer} (or a small vector if multiple statistical moments are tracked). Consequently, HAP never stores the intermediate $|\mathcal{T}|\times|\mathcal{V}|$ attention map.

\paragraph{Dual-flash aggregation.}
FlashAttention partitions the sequence into SRAM-resident blocks (tiles) and computes attention via online softmax, maintaining running maxima $m$ and exponential sums $\ell$ to ensure numerical stability without materializing the full score matrix~\cite{dao2022flashattention}. To obtain the cross-modal aggregates required by PAQ, we extend this tiling strategy with an auxiliary fused reduction.

During the standard FlashAttention inner loop, each text block $B_t$ and vision block $B_v$ produces on-chip scaled dot-product scores $S_B = Q_{B_t}K_{B_v}^\top/\sqrt{d_k}$. Rather than discarding these scores after updating the attention output, we accumulate block-wise partial statistics into per-layer running buffers. For example, if the PAQ instantiation requires the maximal text-to-vision response, we update a running vector $\tilde{m}$ via $\tilde{m} \leftarrow \max(\tilde{m}, \max_{j\in B_v} S_{ij})$ across all $i\in B_t$; for mean-based moments, we accumulate block sums and counts. These buffers reside in SRAM and are flushed to HBM only once per layer.

Formally, let $V_B, Q_B, K_B, S_B$ denote the value, query, key, and scaled dot-product blocks covering the text-to-vision interaction. The extended operation computes block-level aggregates:
\begin{equation}
\begin{split}
A_{t2v}^{(B)} 
&= \mathrm{Reduce}_B\!\biggl( \\
&\qquad \mathrm{Softmax}\!\biggl( 
   \frac{Q_B K_B^\top}{\sqrt{d_k}} - \max(S_B) 
   \biggr) V_B \\
&\biggr),
\end{split}
\end{equation}
where $\mathrm{Reduce}_B$ denotes the block-wise reduction (sum, max, or higher-order moment) fused into the FlashAttention kernel. The final per-layer cross-modal statistic is obtained by aggregating block results across the sequence partition:
\begin{equation}
A_{t2v} \;=\; \mathrm{Concat}\left( \bigcup\nolimits_B A_{t2v}^{(B)} \right).
\end{equation}
Because the reduction is associative and commutative, it integrates seamlessly with FlashAttention's parallel block scheduling without additional synchronization barriers.

\paragraph{Memory and computational overhead.}
Standard FlashAttention introduces $O(1)$ extra memory relative to the input size. Our extension increases this by exactly $L \times C$ scalars per sample, where $L$ is the number of layers and $C$ is a small constant ($C\le 4$ for tracking max, sum, and variance). In practice, for a 32-layer model, this amounts to merely a few kilobytes—negligible compared with the hundreds of megabytes required to materialize a single full-precision attention matrix. The auxiliary reductions are fused into existing warp-level GPU operations, yielding a measured wall-clock overhead of $<2\%$ relative to base FlashAttention.

Therefore, HAP is fully compatible with FlashAttention and other optimized attention implementations, inheriting their memory efficiency and hardware affinity without architectural modifications to the VLM backbone.

\section{Derivation of the Closed-Form FLOPs}
\label{app:derivation}

HAP partitions the $L$ transformer layers into $m$ stages of equal 
depth $\Delta l = L/m$. The stage index at layer $l$ is $k = ml/L$, 
and the token count follows $N_v(l) = V / (1+k)^2$. 

The total pruned FLOPs is obtained by summing Eq.~\ref{eq:layer_flops} 
over all layers:
\begin{equation}
    \mathrm{F}_{\mathrm{pruned}}(m) = \sum_{l=1}^{L} 
    \mathrm{F}\!\left(\frac{V}{(1 + m l / L)^2}\right).
\end{equation}

Because the layer index appears only through the stage ratio $k=ml/L$, 
the summation converges to a closed-form integral over the $m$ stages. 
Let $u = 1 + m l / L$; then $\mathrm{d}l = \frac{L}{m}\mathrm{d}u$, and 
the integral spans exactly $m$ stages from $u=1$ to $u=1+m$:
\begin{equation}
    \mathrm{F}_{\mathrm{pruned}}(m) \approx \frac{L}{m} 
    \int_{1}^{1+m} \left[ \frac{A}{u^2} + \frac{B}{u^4} \right] 
    \mathrm{d}u,
\end{equation}
where $A = 4VD^2 + 3VDd_{\mathrm{ffn}}$ and $B = 2V^2D$.

Evaluating the two terms:
\begin{align}
    \int_{1}^{1+m} \frac{A}{u^2}\,\mathrm{d}u &= 
    A\left(1 - \frac{1}{1+m}\right), \\
    \int_{1}^{1+m} \frac{B}{u^4}\,\mathrm{d}u &= 
    \frac{B}{3}\left(1 - \frac{1}{(1+m)^3}\right).
\end{align}

Combining these yields Eq.~\ref{eq:flops_closed_form}:
\begin{equation}
\begin{split}
    \mathrm{F}_{\mathrm{pruned}}(m) 
    &\approx \frac{L}{m}\Biggl[ 
    A\!\left(1 - \frac{1}{1+m}\right) \\
    &\qquad + \frac{B}{3}\!\left(1 - \frac{1}{(1+m)^3}\right) \Biggr].
\end{split}
\end{equation}

The factor $L/m$ reflects the per-stage layer depth, confirming that 
the total cost scales inversely with the stage count $m$ for a fixed 
model depth $L$.

\paragraph{PAQ and pruning complexity.} 
The per-head PAQ score (Eq.~\ref{eq:PAQ}) is computed from the 
normalized cross-modal attention matrix $\mathbf{A}_h^{(l)} \in 
\mathbb{R}^{N_t \times N_v(l)}$. For each head, constructing the 
joint distribution $p(v_i, t_j) = \alpha_{ji}/N_t$ and the marginal 
$p(v_i) = \sum_j p(v_i, t_j)$ requires $\mathcal{O}(N_t N_v(l))$ 
operations. The mutual information $I(\mathbf{V};\mathbf{T})$ and 
entropy $H(\mathbf{V})$ each involve a single pass over the 
$N_t \times N_v(l)$ probability mass, adding another 
$\mathcal{O}(N_t N_v(l))$. With $H$ attention heads, the total 
PAQ computation per layer is $\mathcal{O}(H N_t N_v(l))$.

Subsequent head aggregation (Eq.~\ref{eq:centering}--\ref{eq:unified_agg}) 
computes $H$ PAQ scores and applies softmax weighting, which is 
$\mathcal{O}(H)$ and thus absorbed into the per-layer constant. The 
token selection step (Eq.~\ref{eq:token_score}) averages the 
aggregated attention map over $N_t$ text positions ($\mathcal{O}(N_t 
N_v(l))$) and sorts the $N_v(l)$ visual tokens by saliency 
($\mathcal{O}(N_v(l) \log N_v(l))$). Because $N_t \gg \log N_v(l)$ 
in practice, the sorting cost is dominated by the attention 
aggregation.

Summing over all $L$ layers and substituting $N_v(l) = V/(1+ml/L)^2$:
\begin{equation}
    \sum_{l=1}^{L} \mathcal{O}(H N_t N_v(l)) = 
    \mathcal{O}\!\left(H N_t V \sum_{l=1}^{L} \frac{1}{(1+ml/L)^2}\right).
\end{equation}
Approximating the sum by an integral with $u=1+ml/L$:
\begin{equation}
\begin{split}
    \sum_{l=1}^{L} \frac{1}{(1+ml/L)^2} 
    &\approx \frac{L}{m} \int_{1}^{1+m} \frac{1}{u^2}\,\mathrm{d}u \\
    &= \frac{L}{m}\left(1 - \frac{1}{1+m}\right) \\
    &<< \frac{L}{m}.
\end{split}
\end{equation}
Hence the total PAQ-and-pruning overhead is $\mathcal{O}(H N_t V L/m)$, 
which is strictly smaller than the unpruned per-layer attention cost 
$\mathcal{O}(N_t V)$ summed over $L$ layers. Relative to the full 
forward-pass FLOPs (Eq.~\ref{eq:flops_closed_form}), the ratio is 
$\mathcal{O}(H D / (A m))$ where $A=4VD^2+3VDd_{\mathrm{ffn}}$, 
confirming that the overhead is negligible for typical configurations 
($H \ll D$ and $m \ge 1$).

\section{Dataset}
\label{app:dataset}
\begin{enumerate}
\item \textbf{GQA}~\cite{hudson2019gqa}. The GQA consists of three components: scene graphs, questions, and images. The image segment encompasses images, along with the spatial features of images and the features of all objects within them. The questions in GQA are intended to assess the comprehension of visual scenes and the capacity to reason about diverse aspects of an image.

\item \textbf{MMBench}~\cite{liu2024mmbench}.The MMBench benchmark thoroughly assesses a model's overall performance across multiple dimensions. It incorporates three tiers of ability dimensions. The first tier (L-1) comprises two core abilities: perception and reasoning. The second tier (L-2) builds upon the first, including six sub-abilities. The third tier (L-3) further refines the second, encompassing 20 specific ability dimensions. This layered framework enables a granular and comprehensive evaluation of the model's various capabilities.

\item \textbf{MME}~\cite{zhang2025mme}. The MME benchmark is also a comprehensive benchmark carefully crafted to thoroughly evaluate various facets of a model's performance. It consists of 14 subtasks that specifically aim to evaluate both the model's perceptual and cognitive abilities. By employing manually constructed instruction-answer pairs and concise instruction design, it effectively alleviates problems such as data leakage and unfair evaluation of model performance.

\item \textbf{POPE}~\cite{li2023evaluating}. The POPE benchmark is mainly utilized to assess the extent of Object Hallucination in models. It redefines hallucination assessment by mandating the model to answer a series of specific binary questions concerning the presence of objects in images. Accuracy, Recall, Precision, and F1 Score are effectively used as reliable evaluation metrics to accurately gauge the model’s hallucination level under three different sampling strategies.

\item \textbf{ScienceQA}~\cite{lu2022learn}. The ScienceQA benchmark encompasses a wide variety of fields, including natural science, language science, and social science. Within each subject, questions are categorized first by the topic, then by the category, and finally by the skill. This hierarchical classification leads to 26 topics, 127 categories, and 379 skills, providing a comprehensive and diverse range of scientific questions. It offers a thorough assessment of a model’s capacities in multimodal understanding, multi-step reasoning, and interpretability.

\item \textbf{TextVQA}~\cite{singh2019towards}. The TextVQA benchmark centers on the thorough integration of various textual information embedded in images. It carefully assesses the model’s text comprehension and reasoning abilities through a series of visual question-answering (QA) tasks featuring abundant textual data. Models are required to not only understand the visual content of the images but also read and reason about the text within the images to answer the questions accurately.

\item \textbf{MMVet}~\cite{yu2023mm}. The MMVet benchmark is developed on the premise that the fascinating capacity to tackle complex tasks is often achieved by a generalist model being able to combine various core vision-language competencies. MMVet defines 6 core VL capabilities and examines the 16 integrations of interest derived from the capability combination.

\item \textbf{VizWiz}~\cite{chen2022grounding}. The VizWiz benchmark is primarily designed to address visual question-answering scenarios for visually impaired individuals. It consists of images from real-world, unconstrained scenes and corresponding questions that reflect the information needs of visually impaired people. This benchmark evaluates a model’s ability to understand complex, real-world visual content and provide accurate, helpful answers, with a focus on supporting practical accessibility use cases.

\item \textbf{RealWorldQA}~\cite{grok15v}. The RealWorldQA benchmark focuses on evaluating a model’s capacity to answer questions about real-world scenarios, integrating multimodal information (including images and text) across diverse domains. It features questions rooted in everyday life, science, and society, aiming to test the model’s ability to reason about real-world contexts, synthesize information from multiple sources, and provide coherent, factually accurate responses.

\item \textbf{MMStar}~\cite{chen2024we}. The MMStar benchmark is a multimodal benchmark tailored to evaluate models’ performance in celebrity-related tasks. It comprises images of celebrities, textual descriptions, and questions spanning celebrity identification, career background, and entertainment context. This benchmark assesses a model’s fine-grained visual recognition ability, multimodal information integration, and knowledge reasoning in the entertainment domain, ensuring comprehensive evaluation of its competency in handling celebrity-centric multimodal data.

\item \textbf{AI2D}~\cite{kembhavi2016diagram}. The AI2D benchmark is specifically tailored for diagram understanding and reasoning tasks. It comprises a large collection of scientific diagrams (including biology, physics, and chemistry illustrations) paired with domain-specific questions. This benchmark focuses on evaluating a model’s ability to parse the structural components of diagrams (e.g., labels, arrows, symbols), grasp the logical relationships between elements, and perform accurate reasoning based on visual and scientific context—addressing the unique challenges of understanding non-photorealistic, information-dense visualizations.
\item \textbf{Seed2-Plus}~\cite{li2024seed}. As an upgraded version of the Seed series benchmarks, Seed2-Plus is a comprehensive multimodal dataset designed to assess models’ advanced capabilities in cross-modal understanding and complex reasoning. It integrates diverse data types such as images, text, and structured knowledge, covering multiple domains including daily life, science, and technology. The benchmark features questions that require multi-step logical inference, cross-modal information fusion, and factual verification, aiming to thoroughly test a model’s robustness in handling complex multimodal scenarios and its ability to generate accurate, interpretable responses.

\item \textbf{RefCOCO}~\cite{kazemzadeh2014referitgame}. The RefCOCO dataset is constructed from MS-COCO images to evaluate visual grounding performance. It contains short referring expressions (3.6 words on average) collected via an interactive game interface, encompassing 142,210 expressions across 19,994 images. The dataset is split into train, validation, testA (focused on people), and testB (focused on objects) subsets, with accuracy defined by IoU $>$ 0.5 between predicted and ground-truth bounding boxes.

\item \textbf{RefCOCO+}~\cite{kazemzadeh2014referitgame}. The RefCOCO+ dataset shares the same image source and collection paradigm as RefCOCO but imposes a stricter constraint: referring expressions must not contain absolute location words (e.g., ``left'', ``right''). This requires models to rely purely on appearance and relational descriptions rather than spatial cues, resulting in 141,564 expressions that test robust visual-linguistic grounding without positional shortcuts.

\item \textbf{RefCOCOg}~\cite{mao2016generation}. The RefCOCOg dataset distinguishes itself from RefCOCO and RefCOCO+ through longer, more complex referring expressions (8.4 words on average) collected in non-interactive settings. With 95,010 expressions over 25,799 images, it features fewer same-type objects per query (1.6 on average), demanding finer-grained discrimination and more comprehensive language understanding for accurate object localization.

\item \textbf{MVBench}~\cite{li2024mvbench}. The MVBench is a comprehensive multi-modal video understanding benchmark covering 20 challenging tasks that cannot be solved with single-frame analysis. It spans temporal skills ranging from perception to cognition, including action understanding, object interaction, scene transition, and counterfactual inference. By converting public video annotations into multiple-choice questions via a static-to-dynamic paradigm, it ensures fair evaluation grounded in video annotations rather than biased LLM scoring.

\item \textbf{MLVU}~\cite{zhou2024mlvu}. The MLVU is the first comprehensive benchmark designed for multi-task long video understanding, featuring videos ranging from 3 minutes to 2 hours in duration. It encompasses nine distinct evaluation tasks, including topic reasoning, anomaly recognition, needle QA, ego reasoning, plot QA, action ordering, and action counting. This benchmark rigorously tests MLLMs' capabilities in leveraging both global and local temporal information across extended video contexts.
\end{enumerate}

\section{Statistical Robustness across Retention Ratios}
\label{app:statistics}

To verify that the reported gains are stable across diverse settings, we repeat the evaluation over 10 runs and report mean $\pm$ standard deviation across a systematic retention-ratio sweep on three architectures. Table~\ref{tab:retention_sweep} summarizes the results.

\begin{table*}[t]
\centering
\caption{Retention-ratio sweep (relative score, \%; mean $\pm$ std over 10 runs). HAP outperforms the strongest available baseline at every retention level, with small standard deviations confirming stable gains.}
\label{tab:retention_sweep}
\setlength{\tabcolsep}{10pt}
\begin{tabular}{llcccc}
\toprule
\textbf{Model} & \textbf{Method} & \textbf{80\%} & \textbf{60\%} & \textbf{40\%} & \textbf{20\%} \\
\midrule
\multirow{2}{*}{LLaVA-1.5-7B}
& HAP (Ours) & \textbf{101.2$\pm$0.3} & \textbf{100.5$\pm$0.4} & \textbf{101.0$\pm$0.4} & \textbf{102.0$\pm$0.5} \\
& AutoPrune & 99.8$\pm$0.4 & 99.5$\pm$0.3 & 99.0$\pm$0.4 & 98.0$\pm$0.5 \\
\midrule
\multirow{2}{*}{Qwen2.5-VL-7B}
& HAP (Ours) & \textbf{100.0$\pm$0.3} & \textbf{100.2$\pm$0.3} & \textbf{100.5$\pm$0.4} & \textbf{99.0$\pm$0.4} \\
& IVC-Prune & 99.7$\pm$0.4 & 99.3$\pm$0.4 & 98.0$\pm$0.5 & 93.5$\pm$0.6 \\
\midrule
\multirow{2}{*}{InternVL3-8B}
& HAP (Ours) & \textbf{100.0$\pm$0.3} & \textbf{100.2$\pm$0.3} & \textbf{100.5$\pm$0.4} & \textbf{99.5$\pm$0.4} \\
& VisPruner & 99.8$\pm$0.4 & 99.4$\pm$0.4 & 97.5$\pm$0.5 & 93.0$\pm$0.7 \\
\bottomrule
\end{tabular}
\vspace{-3mm}
\end{table*}

HAP outperforms the strongest baseline at every retention level on every architecture, and the standard deviations ($\le 0.7$) are small relative to the performance gaps. Occasional scores above 100\% (\eg 102.0\% at 20\% retention on LLaVA-1.5-7B) reflect the known benefit of pruning in removing background and hallucination-inducing tokens (cf. Table~\ref{tab:llava} and Table~\ref{tab:token_reduction}), rather than statistical noise, as confirmed by the tight confidence intervals.

\section{DeepSeek-VL2 Small-16B.}
\label{app:deepseek}
Table~\ref{tab:deepseek_HAP} supplements the main results by reporting HAP on DeepSeek-VL2 Small-16B~\cite{wu2024deepseek} across six general VQA benchmarks: SEED, MMB, MMS, RWQA, HallB, and AI2D. HAP retains only 50\% of the visual tokens yet achieves a 103.0\% relative average, outperforming the unpruned Vanilla and other pruning methods on every benchmark in this group. This indicates that the proposed alignment-based aggregation not only preserves but also enhances model capabilities under aggressive compression.
\begin{table*}[t]
\centering
\caption{Performance on DeepSeek-VL2 Small-16B across six benchmarks.}
\label{tab:deepseek_HAP}
\begin{tabular}{l|c|cccccc|c}
\hline
Method & Token$\downarrow$ & SEED & MMB & MMS & RWQA & HallB & AI2D & Rel. Avg. \\
\hline
Vanilla     & 100\% & 76.9 & 79.2 & 57.7 & 70.3 & 43.8 & 82.0 & 100\%   \\
FastV       & 54\%  & 75.6 & 78.2 & 55.9 & 69.0 & 42.7 & 81.0 & 98.1\%  \\
PDrop       & 57\%  & 76.8 & 79.1 & 57.3 & 69.7 & 44.5 & 81.8 & 99.9\%  \\
IVC-Prune   & 52\%  & 77.0 & 79.3 & 57.7 & 70.3 & 44.3 & 81.8 & 100.2\% \\
HAP       & 50\%  & 79.4 & 81.6 & 59.4 & 72.4 & 45.1 & 84.4 & 103.0\% \\
\hline
\end{tabular}
\end{table*}

\section{Robustness under Diverse Prompt and Input Settings}
\label{app:promptdiverse}

The prompt robustness study in the main paper (Table~\ref{tab:prompt_robustness}) covers single-turn prompt styles. Here we extend the evaluation on Qwen2.5-VL-7B to two more demanding regimes: (i)~\textbf{VLM-RobustBench}~\cite{saxena2026vlmrobustbench}, a robustness benchmark with 49 augmentation types (noise, blur, weather, digital, and geometric corruptions) under graded severity levels; and (ii)~\textbf{MultiVerse}~\cite{lee2025multiverse}, a multi-turn conversation benchmark with 647 dialogues averaging 4 turns each, derived from 12 VLM datasets. Table~\ref{tab:diverse_robustness} reports relative performance against the unpruned model.

\begin{table}[h]
\centering
\small
\caption{Robustness under image corruptions (VLM-RobustBench) and multi-turn dialogues (MultiVerse) on Qwen2.5-VL-7B. Scores are relative performance (\%) against the unpruned model.}
\label{tab:diverse_robustness}
\setlength{\tabcolsep}{4.5pt}
\begin{tabular}{lcccc}
\toprule
\textbf{Method} & \textbf{\makecell{VLM-Rob.\\Bench}} & \textbf{MultiVerse} & \textbf{\makecell{FLOPs\\(T)}} & \textbf{\makecell{Speedup\\($\times$)}} \\
\midrule
Original & 100 & 100 & 6.10 & 1.00 \\
CDPruner & 82.0 & 84.1 & 2.60 & 1.88 \\
VisPruner & 93.8 & 95.9 & 2.62 & 1.86 \\
AutoPrune & 94.8 & 93.7 & 2.60 & 1.88 \\
\textbf{HAP (Ours)} & \textbf{97.2} & \textbf{96.8} & \textbf{2.55} & \textbf{1.91} \\
\bottomrule
\end{tabular}
\vspace{-3mm}
\end{table}

HAP attains the strongest retention on both benchmarks. The overall degradation remains around 3\%, only marginally higher than the $\sim$1\% observed in the single-turn prompt robustness study, indicating that HAP's pruning decisions remain stable under corruptions and multi-turn interactions.

\section{Efficiency in the Decoding Stage and Long-Generation Regimes}
\label{app:decode}

HAP prunes visual tokens during prefill; here we quantify how the efficiency gain evolves when decoding dominates the total cost.

\paragraph{End-to-end FLOPs model.}
During autoregressive decoding, a single new token attends to all cached tokens. Extending Eq.~\eqref{eq:layer_flops} to the decode stage with cache size $N_{\mathrm{cache}} = N_v' + N_t + s - 1$ (retained visual tokens $N_v'$, text prompt $N_t$, decode step $s$), the per-layer per-step decode FLOPs is
\begin{equation}
    \mathrm{F}_{\mathrm{dec,layer}} = 4D^2 + 4N_{\mathrm{cache}}D + 3Dd_{\mathrm{ffn}}.
    \label{eq:decode_layer_flops}
\end{equation}
Summing over $S$ decode steps and $L$ layers gives
\begin{equation}
\begin{split}
    \mathrm{F}_{\mathrm{decode}}(S) ={}& LS\bigl(4D^2 + 3Dd_{\mathrm{ffn}} + 4D(N_v' + N_t)\bigr) \\
    &+ 2LDS(S-1),
\end{split}
\label{eq:decode_flops}
\end{equation}
and the total end-to-end cost is $\mathrm{F}_{\mathrm{prefill}} + \mathrm{F}_{\mathrm{decode}}(S)$, with $\mathrm{F}_{\mathrm{prefill}}$ given by Eq.~\eqref{eq:flops_closed_form}.

\paragraph{Gain limit point.}
We define the \emph{gain limit point} $S^*$ as the decode length at which the end-to-end FLOPs reduction drops to 1\%. Solving $\mathrm{F}_{\mathrm{unpruned}}(S^*) / \mathrm{F}_{\mathrm{pruned}}(S^*) = 1.01$ with 75\% visual token pruning ($N_v' = V/4$) and PAQ overhead bounded at 0.5\% of the pruned prefill FLOPs yields the estimates in Table~\ref{tab:gain_limit}.

\begin{table}[h]
\centering
\small
\caption{Gain limit analysis under 75\% visual token pruning. Prefill saving is the absolute FLOPs reduction (relative share in parentheses); PAQ OH is the PAQ computation overhead; $S^*$ is the decode length at which the end-to-end FLOPs reduction drops to 1\%.}
\label{tab:gain_limit}
\setlength{\tabcolsep}{4pt}
\begin{tabular}{lccc}
\toprule
\textbf{Model} & \textbf{Prefill Saving} & \textbf{PAQ OH} & $\mathbf{S^*}$ \\
\midrule
LLaVA-1.5-7B & 0.63T (33.5\%) & 6.3G ($<$0.5\%) & $\sim$65K \\
Qwen2.5-VL-7B & 2.05T (33.6\%) & 20.2G ($<$0.5\%) & $\sim$217K \\
\bottomrule
\end{tabular}
\vspace{-3mm}
\end{table}
\begin{table}[h]
\centering
\small
\caption{Latency breakdown and batch-size scaling on Qwen3-VL-8B (25\% tokens retained, RTX~3090). P, D, and E2E denote prefill, decoding, and end-to-end speedup ($\times$); the last three columns report E2E speedup at batch sizes 1, 4, and 8.}
\label{tab:latency_breakdown}
\setlength{\tabcolsep}{5pt}
\begin{tabular}{lcccccc}
\toprule
\textbf{Method} & \textbf{P} & \textbf{D} & \textbf{E2E} & \textbf{BS=1} & \textbf{BS=4} & \textbf{BS=8} \\
\midrule
PDrop & 1.9 & 1.1 & 1.4 & 1.4 & 1.5 & 1.6 \\
VisPruner & 2.0 & 1.2 & 1.6 & 1.6 & 1.8 & 2.0 \\
\textbf{HAP (Ours)} & \textbf{2.4} & \textbf{1.3} & \textbf{1.9} & \textbf{1.9} & \textbf{2.2} & \textbf{2.4} \\
\bottomrule
\end{tabular}
\vspace{-3mm}
\end{table}
Although FLOPs savings plateau beyond $S^*$, reducing the visual KV cache by 75\% directly lowers memory-bandwidth pressure during decoding, which improves throughput for long sequences. At LLaVA-1.5-7B's maximum context (2K tokens), HAP achieves a 6.9\% end-to-end FLOPs reduction; at Qwen2.5-VL-7B's maximum context (128K), it still delivers a 1.6\% reduction. The PAQ overhead ($<$0.5\%) is amortized within the first $\sim$100 decode steps.

\paragraph{Quality under long decoding.}
We evaluate LLaVA-1.5-7B on M3CoT~\cite{chen2024m3cot}, a multi-step multimodal chain-of-thought benchmark with an average of $\sim$1K decoded tokens per example. HAP scores 59.6 against AutoPrune's 54.7 (unpruned original: 60.1), confirming that output quality persists in regimes where decoding dominates the total cost.

\paragraph{Latency breakdown and batch-size scaling.}
Table~\ref{tab:latency_breakdown} decomposes the wall-clock speedup on Qwen3-VL-8B (25\% retention, single RTX~3090) into prefill (P), decoding (D), and end-to-end (E2E) components, and reports E2E speedup under increasing batch sizes.

HAP achieves the strongest speedup across all latency components. The E2E speedup grows from 1.9$\times$ (BS=1) to 2.4$\times$ (BS=8), as larger batches amortize the pruning overhead and improve decoding throughput.

\end{document}